\documentclass[a4paper,conference]{IEEEtran}
\usepackage{cite}

\ifCLASSINFOpdf
\else
\fi
\usepackage{amsmath}
\usepackage{amssymb}
\newcommand{\spm}[1]{{\tiny$\pm$#1}}

\ifCLASSOPTIONcompsoc
  \usepackage[caption=false,font=normalsize,labelfont=sf,textfont=sf]{subfig}
\else
  \usepackage[caption=false,font=footnotesize]{subfig}
\fi
\usepackage{dblfloatfix}

\usepackage{url}

\usepackage{pgfplots}
\usepackage{booktabs}
\usepackage{tikz}
\usepackage{algorithm}
\usepackage{algpseudocode}
\usepackage{dsfont}
\usetikzlibrary{positioning}
\usepackage{multirow}
\pgfplotsset{width=10cm,compat=1.9}
\usepgfplotslibrary{groupplots}

\begin{document}
%
\title{DoubleMatch:\\ Improving Semi-Supervised Learning with Self-Supervision}

\author{\IEEEauthorblockN{Erik Wallin\textsuperscript{1,2}, Lennart Svensson\textsuperscript{2}, Fredrik Kahl\textsuperscript{2}, Lars Hammarstrand\textsuperscript{2}}
\IEEEauthorblockA{\textsuperscript{1}Saab AB, \textsuperscript{2}Chalmers University of Technology\\
\{walline,lennart.svensson,fredrik.kahl,lars.hammarstrand\}@chalmers.se}}


%


\maketitle

\begin{abstract}

Following the success of supervised learning, semi-supervised learning (SSL) is now becoming increasingly popular. SSL is a family of methods, which in addition to a labeled training set, also use a sizable collection of unlabeled data for fitting a model. Most of the recent successful SSL methods are based on pseudo-labeling approaches: letting confident model predictions act as training labels. While these methods have shown impressive results on many benchmark datasets, a drawback of this approach is that not all unlabeled data are used during training. We propose a new SSL algorithm, DoubleMatch, which combines the pseudo-labeling technique with a self-supervised loss, enabling the model to utilize all unlabeled data in the training process. We show that this method achieves state-of-the-art accuracies on multiple benchmark datasets while also reducing training times compared to existing SSL methods. Code is available at \url{https://github.com/walline/doublematch}.

\end{abstract}


%
\IEEEpeerreviewmaketitle

\section{Introduction}

Supervised learning has gained much attention in recent years because of remarkable achievements in fields such as image classification \cite{dosovitskiy2020image}, object detection \cite{redmon2016you}, and natural language processing \cite{vaswani2017attention}. The impressive results are typically fueled by vast amounts of labeled data with datasets such as ImageNet \cite{deng2009imagenet} and COCO \cite{lin2014microsoft}. In many practical applications, however, labeled data might be scarce or require expert domain knowledge to attain. In contrast, \emph{unlabeled data} are often much easier to acquire, e.g., through web scraping or unsupervised sensor recordings, thus creating a natural demand for methods that can successfully learn from data \emph{without} prior labels.

Semi-supervised learning is the procedure of combining unlabeled data with a (typically) much smaller set of labeled data for fitting a model. There are two main ideas behind the most well-performing methods for semi-supervised learning: \emph{consistency regularization} and \emph{pseudo-labeling}. The former means encouraging consistent predictions on unlabeled data across different views of the same sample, e.g., through domain-specific data augmentation. Pseudo-labeling, on the other hand, involves using confident model predictions on unlabeled data as de facto training labels.

UDA \cite{xie2019unsupervised} and FixMatch \cite{sohn2020fixmatch} recently gained wide recognition because of their simple yet powerful frameworks for combining consistency regularization and pseudo-labeling in semi-supervised learning. Among their reported results is, e.g.,\ an impressive classification accuracy of 94.93\% on CIFAR-10 \cite{krizhevsky2009learning} using only 250 labeled images for training \cite{sohn2020fixmatch}. However, despite their excellent performance, these methods have the drawback of only enforcing consistency regularization on unlabeled data with confident model predictions while harder data samples are essentially discarded. 
For challenging datasets, in particular, this means the model only uses a smaller part of unlabeled data during training (the relatively easier part). We show that this inefficient and incomplete use of unlabeled data leads to unnecessarily long training times and, for some datasets, reductions in classification accuracy.



\begin{figure}[]
    \centering
    \footnotesize
    \includegraphics[width=0.75\columnwidth]{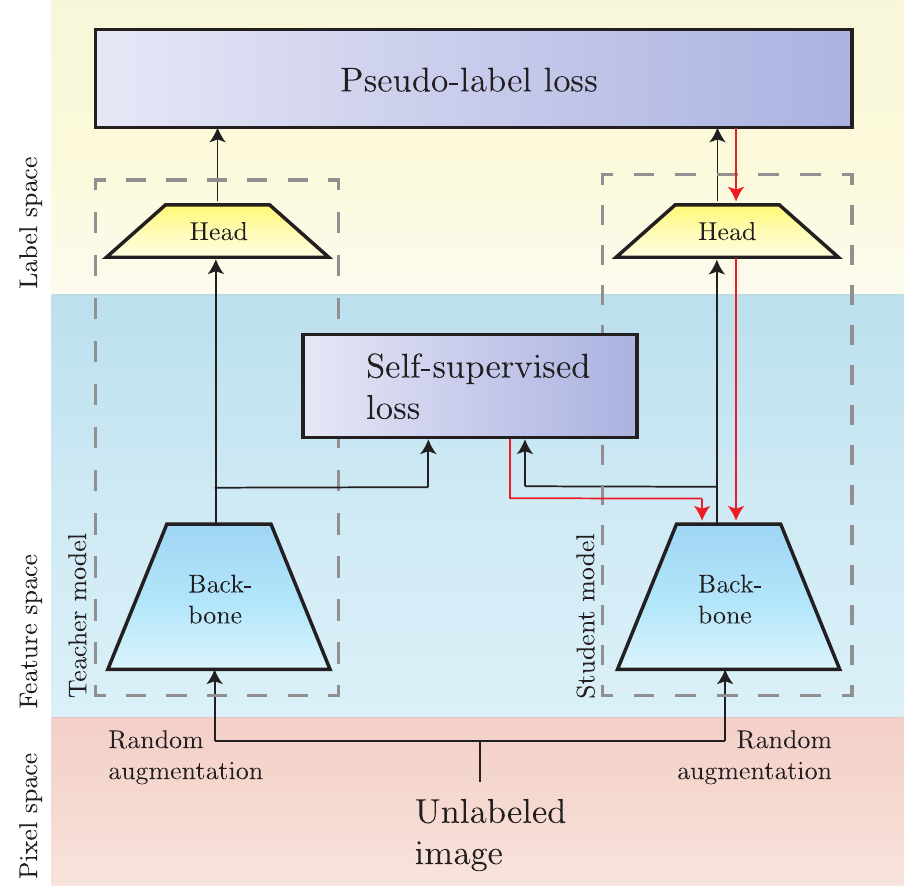}
    \caption{A unification of semi- and self-supervised frameworks. While many existing  methods for semi-supervised learning operate in the label space, DoubleMatch operates in both the label and feature spaces for a more efficient use of unlabeled data. Red arrows mark the flow of gradients.}
    \label{fig:intro}
    \vspace*{-10pt}
\end{figure}


An alternative to semi-supervised training is to instead train fully without label information. This practice is referred to as self-supervised learning, and the goal is not to predict classes but instead to learn feature representations of the data that can be used for downstream tasks \cite{he2020momentum, chen2020simple, chen2021exploring}. By focusing on learning the fundamental representations of the data, these methods can more fully exploit the unlabeled training data and are in this respect more efficient than many semi-supervised methods. However, if the end task is classification, these methods do not extend the training data with task-specific information as in the case of, e.g., UDA and FixMatch. 

Analogously to semi-supervised learning, self-supervised learning is heavily based on consistency regularization, which makes the two fields closely related. Generally, methods for both semi- and self-supervised learning employ a setting where a \emph{teacher model} predicts optimization targets for a \emph{student model}, making gradients flow only through the student predictions. The teacher and student models often share parameters, either by making the teacher a moving average of the student or by using the same parameters. 


We propose DoubleMatch, an extension of FixMatch, which solves its inefficient use of unlabeled data by taking inspiration from research on self-supervised learning. Fig.~\ref{fig:intro} illustrates how DoubleMatch unifies the typical setups for semi- and self-supervised learning by operating both in the label and feature space for unlabeled data. In the label space, as with FixMatch for confident data, the model is evaluated based on predicted class distributions (\emph{pseudo-label loss}). Whereas, in the feature space, the model is assessed based on the similarity of the predicted feature representations (\emph{self-supervised loss}). 
More specifically, with motivation from methods such as \cite{chen2021exploring, grill2020bootstrap}, we suggest adding a self-supervised feature loss to the FixMatch framework by enforcing consistency regularization on \emph{all} unlabeled data in the feature space. Moreover, we can add this term with minimal computational overhead because we do not require additional augmentations or model predictions.

Our main contributions of this work are:
\begin{enumerate}
    \item We propose adding a self-supervised loss to the FixMatch framework.
    \item We demonstrate that our method leads to faster training times and increased classification accuracy than previous SOTA across multiple datasets.
    \item We analyze the importance of a pseudo-labeling loss for different sizes of the labeled training set.
\end{enumerate}


\section{Related work}


Research in semi-supervised learning dates back to the 1960s \cite{scudder1965probability, fralick1967learning, agrawala1970learning}. Over the last couple of years, research in semi-supervised learning has been dominated by methods combining the use of consistency regularization and pseudo-labeling \cite{sohn2020fixmatch, xie2019unsupervised}. One of these methods is FixMatch \cite{sohn2020fixmatch}, which this work proposes an extension of.

\subsection{FixMatch}

FixMatch \cite{sohn2020fixmatch} made a big impact on the field of semi-supervised learning because of its well-performing yet simple method for combining consistency regularization and pseudo-labeling. FixMatch is similar to the marginally earlier UDA \cite{xie2019unsupervised} although slightly less complex. 

As proposed by ReMixMatch \cite{berthelot2019remixmatch}, FixMatch employs consistency regularization through a setup with weak and strong data augmentations. The weak augmentation consists of a random horizontal flip followed by a random image translation, while the strong augmentation utilizes sharp image-transformations such as RandAugment \cite{cubuk2020randaugment}, CTAugment \cite{berthelot2019remixmatch}, and CutOut \cite{devries2017improved}. The strong domain-specific augmentations for consistency regularization are highlighted as a crucial component in the performance gain compared to many previous methods, which use, e.g., domain-agnostic perturbations \cite{rasmus2015semi, laine2016temporal, tarvainen2017mean, miyato2018virtual} or MixUp-augmentations \cite{zhang2017mixup, berthelot2019mixmatch, verma2019ict}.

FixMatch also makes use of pseudo-labeling, which can be compactly described as the practice of using model predictions as training labels. While there are many variations of this technique \cite{yarowsky1995unsupervised, lee2013pseudo, pham2021meta, rizve2021defense}, FixMatch follows the confidence-based pseudo-labels introduced by UDA. The confidence-based pseudo-labels in FixMatch are designed such that only predictions on unlabeled data above a predetermined confidence-threshold are used for training. The artificial label used in the cross-entropy loss is then the argmax of the confident prediction.

The proposed method in this article is an extension of the FixMatch framework, following both its augmentation scheme and pseudo-labeling strategy. 

\subsection{Extensions of FixMatch}

As a consequence of the simplicity and high performance of FixMatch, there have been several subsequent studies trying to extend and improve on this framework \cite{zhang2021flexmatch, xu2021dash, xu2021dp, hu2021simple, nassar2021all}. For example, FlexMatch \cite{zhang2021flexmatch} and Dash \cite{xu2021dash} propose two different ways of introducing dynamic confidence-thresholds, replacing the constant threshold of FixMatch. DP-SSL \cite{xu2021dp} generates pseudo-labels using a data programming scheme with an ensemble of labeling functions, each specialized in a subset of the classes in the classification problem. In contrast to our work, methods such as \cite{zhang2021flexmatch, xu2021dash, xu2021dp, nassar2021all} aim to improve FixMatch through alternative strategies for pseudo-label selections. Our proposed method instead focuses on improving the consistency regularization of FixMatch and thus could potentially be further improved by using pseudo-labeling strategies proposed by \cite{zhang2021flexmatch, xu2021dash, xu2021dp, nassar2021all}.



\begin{figure*}[]
    \centering
    \includegraphics[width  =0.8\linewidth]{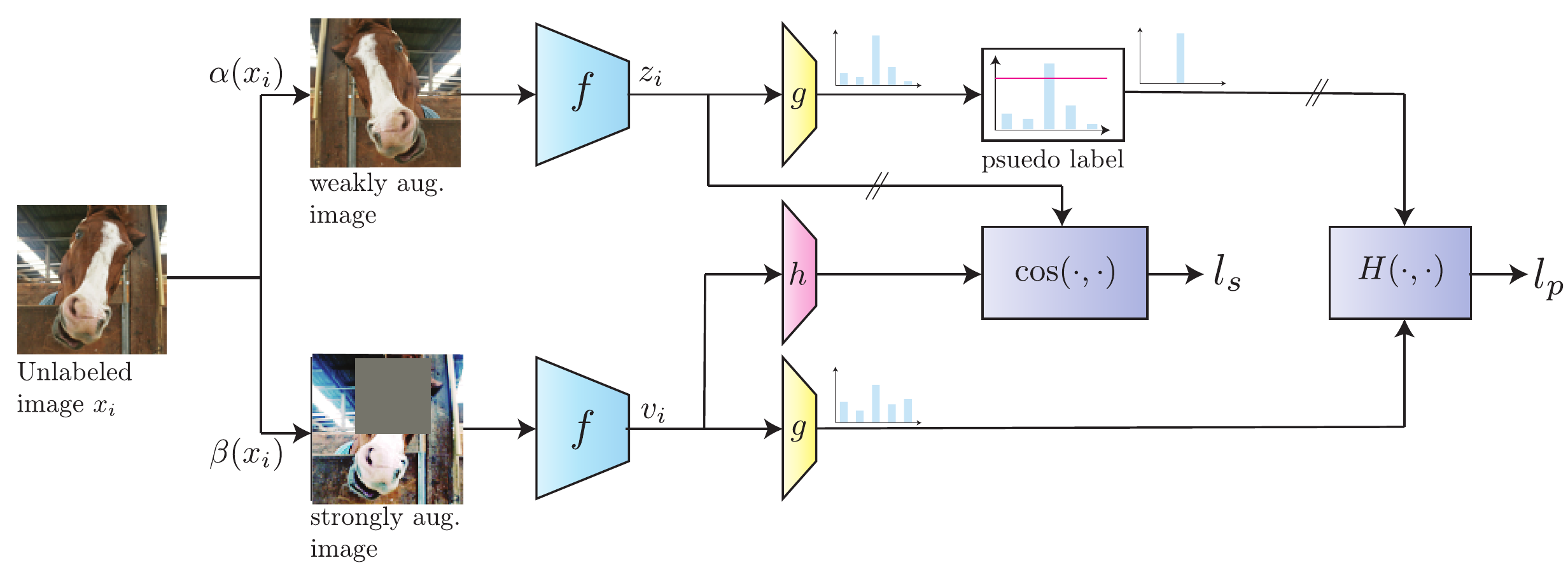}
    \caption{Graph showing loss evaluation for unlabeled images. Double slash marks a stop-gradient operation.}
    \label{fig:uloss}
\end{figure*}

\subsection{Self-supervised learning}

This section gives a very brief overview of research in self-supervised learning that has inspired our work. Fully without labels, the goal of self-supervised learning is not to predict classes, but instead to learn \emph{representations} of data that can be used for downstream tasks. Similarly to semi-supervised learning, recent advances in self-supervised learning have been largely reliant on consistency regularization through heavy use of domain-specific data augmentation \cite{he2020momentum, chen2020simple, grill2020bootstrap, chen2021exploring, caron2021emerging}.

Influential works such as \cite{he2020momentum, chen2020simple} make use of the compute-heavy contrastive loss for self-supervised learning, not only maximizing similarities between augmentations of the same data but also \emph{minimizing} the agreement between different data. However, more recent works \cite{grill2020bootstrap, chen2021exploring, caron2021emerging} show that a contrastive loss is not required to achieve competitive self-learning performance. Instead, they show that it is sufficient to, as in, e.g., FixMatch, focus on maximizing similarities between augmentations of the same data (consistency regularization). This means that self-supervised learning losses similar to those in \cite{grill2020bootstrap, chen2021exploring, caron2021emerging} can be integrated into, e.g., FixMatch with a minimal additional computational load. For this work, we take particular inspiration from SimSiam \cite{chen2021exploring} by incorporating a similar self-supervised loss into our method.


\subsection{Self-supervision in semi-supervised learning}

Lastly, similarly to this paper, there has been some other work covering self-supervision techniques in semi-supervised learning \cite{wang2020enaet, berthelot2019remixmatch, zhai2019s4l, han2020unsupervised, chen2020big}. For example, EnAET \cite{wang2020enaet}, ReMixMatch \cite{berthelot2019remixmatch}, and S4L \cite{zhai2019s4l} incorporate a self-supervised task that involves predicting the parameters of some stochastic image transformation that is applied to an unlabeled image. USADTM \cite{han2020unsupervised} proposes a self-supervised loss that involves evaluating similarities between \emph{three} different views of unlabeled data.
Contrary to our proposed method, the self-supervision in these methods 
requires introducing additional data augmentations and, as in case of USADTM, a more complex pseudo-label selection, all of which increases the computational load. 


\section{Method}

This section describes DoubleMatch, our proposed method. Many existing methods for semi-supervised learning have two central terms in their loss functions: a supervised loss term for labeled data and a pseudo-label loss term for unlabeled data \cite{sohn2020fixmatch, xie2019unsupervised, lee2013pseudo}. We propose to add a third, self-supervised loss term, to fully utilize unlabeled data for faster training and increased accuracy. This term acts as a natural extension with minimal computational overhead to methods that utilize consistency regularization through data augmentation. We choose to apply our idea to the FixMatch \cite{sohn2020fixmatch} framework but the method proposed can easily be applied to similar frameworks such as UDA \cite{xie2019unsupervised}. The training algorithm of DoubleMatch for a single batch is summarized in Algorithm \ref{alg:doublematch} and detailed below.

For a $C$-class image classification problem, the supervised loss in FixMatch is the standard cross-entropy loss given by
\begin{equation}
    l_l = \frac{1}{B} \sum_{i=1}^B H(y_i, p_i),
    \label{eq:supervised-loss}
\end{equation}
where $B$ is the batch size, $y_i \in \mathbb{R}^{C \times 1}$ is the ground-truth one-hot label for image $i$, $p_i \in \mathbb{R}^{C \times 1}$ is the predicted probability distribution for image $i$, and $H(x,y) = -\sum x \log y$ is the cross-entropy between two probability distributions.

The supervised loss is supplemented with a pseudo-label loss on unlabeled data: 
\begin{equation}
    l_p = \frac{1}{\mu B} \sum_{i=1}^{\mu B} \mathds{1} \{ \max(w_i) > \tau \} H(\text{argmax}(w_i), q_i).
    \label{eq:pseudo-label-loss}
\end{equation}
Here, $\mu \in \mathbb{Z}^+$ and $\tau \in [0,1]$ are hyperparameters where $\mu$ determines the ratio of labeled to unlabeled data in a batch and $\tau$ sets the confidence threshold for assigning pseudo-labels. The probability distributions $w_i$ and $q_i$ are the predictions for a \emph{weak} and a \emph{strong} augmentation, respectively, of the unlabeled sample $i$. We let $\text{argmax}: \mathbb{R}^{C \times 1} \rightarrow \{0,1\}^{C \times 1}$ so that it returns a one-hot vector, and $\mathds{1}\{\cdot\}$ is the indicator function. Finally, we let the prediction on the weakly augmented image act as the teacher, meaning we consider $w_i$ to be constant when back-propagating through this loss term.

The two loss terms from \eqref{eq:supervised-loss} and \eqref{eq:pseudo-label-loss} constitute the key components of the FixMatch method. In our method, we add an additional consistency regularization between the two different augmentations of unlabeled data. This consistency regularization is enforced not on the predicted probability distributions but on the \emph{feature vectors} from the penultimate layer of the classification network. This added loss improves the data efficiency of the method by operating on all unlabeled data, and not only for data with confident predictions as in \eqref{eq:pseudo-label-loss}. Inspired by the SimSiam \cite{chen2021exploring} method for self-supervised learning, we use the negative cosine similarity for this loss. Similarly to \cite{chen2021exploring}, in order to allow for differences between the feature representations of weakly and strongly augmented versions of the same image, we feed the feature representation of the strongly augmented image through a trainable linear \emph{projection head} before evaluating the cosine similarity. We end up with our proposed additional loss term:
\begin{equation}
    l_s = - \frac{1}{\mu B} \sum_{i=1}^{\mu B} \frac{h(v_i) \cdot z_i}{\|h(v_i)\| \|z_i\|} = -\frac{1}{\mu B} \sum_{i=1}^{\mu B} \cos (h(v_i), z_i).
    \label{eq:cosine}
\end{equation}
In this expression, $z_i \in \mathbb{R}^{d \times 1}$ is the output from the penultimate layer of the classification network for the weakly augmented version of image $i$. The dimension of this vector, $d$, is determined by the network architecture. Similarly, $v_i \in \mathbb{R}^{d \times 1}$ is the output from the penultimate layer for the strongly augmented version of image $i$. However, $v_i$ is also fed through the trainable linear projection head, $h: \mathbb{R}^{d \times 1} \rightarrow \mathbb{R}^{d \times 1}$. Again, the prediction on the weakly augmented image acts as the teacher, so we consider $z_i$ as constant when evaluating the gradient w.r.t. this loss term. Finally, $\|\cdot\|$ is the $l^2$ norm.

For each batch, we let the full loss be
\begin{equation}
    l = l_l + l_p + w_s l_s,
\end{equation}
where $w_s \in \mathbb{R}^+$ is a hyperparameter determining the weight of the self-supervised loss term. We empirically find that it is important to obtain a good balance between $l_l$ and $w_s l_s$ in this loss, meaning that a well-tuned value for $w_s$ will be largely correlated with the number of labeled training data (see Section \ref{sec:hyper}). Note that our loss function is identical to that used in FixMatch when $w_s=0$. A graph showing the loss evaluation for unlabeled images is displayed in Fig. \ref{fig:uloss} where we divide the network into three parts:
\begin{itemize}
    \item $f$, backbone: the network up to the final layer,
    \item $g$, prediction head: the final layer of the classification network,
    \item $h$, projection head: a single dimension-preserving linear layer for transforming features of strongly augmented unlabeled images.
\end{itemize}

\subsection{Data augmentation}

Data augmentation has proved to be a central component of self- and semi-supervised learning \cite{sohn2020fixmatch, chen2020simple}. We follow one of the augmentation schemes used in FixMatch \cite{sohn2020fixmatch} where the weak augmentation is a horizontal flip with probability 0.5 followed by a random translation with maximum distance 0.125 of the image height. The strong augmentation stacks the weak augmentation, CTA \cite{berthelot2019remixmatch}, and Cutout \cite{devries2017improved}.

\subsection{Optimizer and regularization}

For optimization, we stay close to the FixMatch settings and use SGD with Nesterov momentum \cite{nesterov1983method}. The learning rate, $\eta$, is set to follow a cosine scheme \cite{loshchilov2016sgdr} given by
\begin{equation}
    \eta = \eta_0 \cos \left( \gamma \frac{\pi k}{2K} \right)
\end{equation}
where $\eta_0$ is the initial learning rate, $k$ is the current training step and $K$ is the total number of training steps. We define one training step as one gradient update in the SGD optimization. The decay rate is determined by the hyperparameter $\gamma \in (0,1)$. Contrary to FixMatch which uses a constant $\gamma$, we suggest tuning  $\gamma$ for different datasets in order to minimize overfitting.

Finally, we add weight-decay regularization to the loss as
\begin{equation}
    l_w = w_d \frac{1}{2} \left( \|\theta_f\|^2 + \|\theta_g\|^2 + \|\theta_h\|^2 \right),
\end{equation}
where $\theta_f$, $\theta_g$ and $\theta_h$ are the vectors of parameters in the backbone, prediction head and projection head, respectively, and $w_d$ is a hyperparameter controlling the weight of this regularization term. The weight-decay is identical to FixMatch with the exception that DoubleMatch has the additional parameters from the projection head, $\theta_h$.



\begin{algorithm}[t]
    \centering
    \footnotesize
    \algnewcommand{\LineComment}[1]{\State \(\triangleright\) #1}
\begin{algorithmic}[1]
    \Require Strong augmentation $\beta$, weak augmentation $\alpha$, labeled batch $\{(x_1, y_1),\cdots,(x_B,y_B)\}$, unlabeled batch $\{\tilde{x}_1,\cdots,\tilde{x}_{\mu B}\}$, unsupervised loss weight $w_s$, weight decay parameter $w_d$, confidence threshold $\tau$, backbone model $f$, prediction layer $g$, projection layer $h$
    
    \Statex
    \LineComment{Cross-entropy loss for (weakly augmented) labeled data}
    \For{$i = 1,\cdots, B$} 
        \State $p_i = g \circ f (\alpha(x_i))$
    \EndFor
    \State $l_l = \frac{1}{B} \sum_{i=1}^B H(y_i,p_i)$
    \Statex
    \LineComment Predictions on unlabeled data
    \For{$i = 1, \cdots, \mu B$} 
        \State $z_i = f(\alpha(\tilde{x}_i))$ \Comment{Weak augmentation}
        \State $v_i = f(\beta(\tilde{x}_i))$ \Comment{Strong augmentation}
        \State $q_i = g(v_i)$
        \State $w_i = \texttt{stopgrad}(g(z_i))$
    \EndFor
    
    \LineComment{Self-supervised cosine similarity} 
    \State $l_s = -\frac{1}{\mu B} \sum_{i=1}^{\mu B} \cos(h(v_i), \texttt{stopgrad}(z_i))$
    \LineComment{Cross-entropy with pseudo-labels}
    \State $l_p = \frac{1}{\mu B} \sum_{i=1}^{\mu B} \mathds{1}\{ \max(w_i) > \tau\} H(\text{argmax}(w_i), q_i)$ \vspace{1em}
    \Statex
    \Return $l_l + l_p + w_s l_s + w_d \frac{1}{2}\left( \|\theta_f\|^2 + \|\theta_g\|^2 + \|\theta_h\|^2 \right)$
\end{algorithmic}
    \caption{DoubleMatch algorithm}
    \label{alg:doublematch}
\end{algorithm}

\begin{table*}[]
    \footnotesize
    \centering
    \setlength{\tabcolsep}{2.0pt}
    \caption{Error rates on different datasets using different sizes for the labeled training set. DoubleMatch achieves state-of-the-art results on many combinations.}
    \label{tab:accuracies}
    \begin{tabular}{c c c c c c c c c c c}
\toprule
 & \multicolumn{3}{c}{CIFAR-10} & \multicolumn{3}{c}{CIFAR-100} & \multicolumn{3}{c}{SVHN} & STL-10 \\ \cmidrule(lr){2-4} \cmidrule(lr){5-7} \cmidrule(lr){8-10} \cmidrule(lr){11-11} 
 Method & 40 labels & 250 labels & 4000 labels & 400 labels & 2500 labels & 10000 labels & 40 labels & 250 labels & 1000 labels & 1000 labels \\ \midrule
 MixMatch \cite{sohn2020fixmatch, berthelot2019mixmatch} & 47.54\spm{11.50} & 11.05\spm{0.86} & 6.42\spm{0.10} & 67.61\spm{1.32} & 39.94\spm{0.37} & 28.31\spm{0.33} & 42.55\spm{14.53} & 3.98\spm{0.23} & 3.50\spm{0.28} & 10.41\spm{0.61} \\
 UDA \cite{sohn2020fixmatch, xie2019unsupervised} & 29.05\spm{5.93} & 8.82\spm{1.08} & 4.88\spm{0.18} & 59.28\spm{0.88} & 33.13\spm{0.22} & 24.50\spm{0.25} & 52.63\spm{20.51} & 5.69\spm{2.76} & 2.46\spm{0.24} & 7.66\spm{0.56} \\
 ReMixMatch \cite{sohn2020fixmatch, berthelot2019remixmatch} & 19.10\spm{9.64} & 5.44\spm{0.05} & 4.72\spm{0.13} & 44.28\spm{2.06} & 27.43\spm{0.31} & 23.03\spm{0.56} & 3.34\spm{0.20} & 2.92\spm{0.48} & 2.65\spm{0.08} & 5.23\spm{0.45} \\
 FixMatch (CTA) \cite{sohn2020fixmatch} & 11.39\spm{3.35} & 5.07\spm{0.33} & 4.31\spm{0.15} & 49.95\spm{3.01} & 28.64\spm{0.24} & 23.18\spm{0.11} & 7.65\spm{7.65} & 2.64\spm{0.64} & 2.36\spm{0.19} & 5.17\spm{0.63} \\
 DP-SSL \cite{xu2021dp} & {\bf 6.54\spm{0.98}} & {\bf 4.78\spm{0.26}} & 4.23\spm{0.20} & 43.17\spm{1.29} & 28.00\spm{0.79} & 22.24\spm{0.31} & {\bf 2.98\spm{0.86}} &  {\bf 2.16\spm{0.36}} & {\bf 1.99\spm{0.18}} & 4.97\spm{0.42} \\
 Dash (CTA) \cite{xu2021dash} & 9.16\spm{4.31} & {\bf 4.78\spm{0.12}} & {\bf 4.13\spm{0.06}} & 44.83\spm{1.36} & 27.85\spm{0.19} & 22.77\spm{0.21} & 3.14\spm{1.60} & 2.38\spm{0.29} & 2.14\spm{0.09} & {\bf 3.96\spm{0.25}} \\ \midrule
 DoubleMatch (last 20) & 14.02\spm{5.71} & 5.72\spm{0.51} & 4.83\spm{0.17} & {\bf 42.61\spm{1.15}} & {\bf 27.48\spm{0.19}} & {\bf 21.69\spm{0.26}} & 16.50\spm{13.73} & 2.58\spm{0.53} & 2.25\spm{0.09} & 4.46\spm{0.20} \\
 DoubleMatch (min) & 13.59\spm{5.60} & 5.56\spm{0.42} & 4.65\spm{0.17} & {\bf 41.83\spm{1.22}} & {\bf 27.07\spm{0.26}} & {\bf 21.22\spm{0.17}} & 15.37\spm{11.81} & 2.37\spm{0.35} & 2.10\spm{0.07} & 4.35\spm{0.20} \\\bottomrule
 
\end{tabular}
\end{table*}

\section{Experiments/results}

In this section we evaluate our method on a set of benchmark datasets for image classification. We use the datasets CIFAR-10 \cite{krizhevsky2009learning}, CIFAR-100 \cite{krizhevsky2009learning}, SVHN \cite{netzer2011reading} and STL-10 \cite{coates2011analysis} with different sizes of the labeled training set. We compare our results with reported error rates from similar works. For fair comparisons, we choose works that state using similar experimental setups as us with the same data folds and the same architectures \cite{sohn2020fixmatch, xu2021dash, xu2021dp}. The results are shown in Table \ref{tab:accuracies} where we present mean and standard deviation of the error rate on the test set using five different data folds. We choose to report both the minimum error rate for the full training run, and the median for the 20 last evaluations. FixMatch \cite{sohn2020fixmatch} uses the median of the last evaluations for their results, while others are not clear on this point.

We use an exponential moving average of the model parameters (with momentum 0.999) to evaluate the performance on the test set. We train DoubleMatch using $352,000$ training steps with batch size $B = 64$ in all our experiments. The method is implemented in TensorFlow and is based on the FixMatch codebase. All experiments are carried out on Nvidia A100 GPUs. The full list of hyperparameters can be found in Section \ref{sec:hyper}.

\subsection{Classification results}

\subsubsection{CIFAR-10 and SVHN}

The datasets CIFAR-10 and SVHN both consist of color images of size $32\times32$. Both datasets contain ten classes where CIFAR-10 has classes such as dog, horse and ship while the classes in SVHN are the ten digits. CIFAR-10 has a test set of 10,000 images and a training set of 50,000 images. SVHN has 26,032 images for testing and 73,257 for training. For these datasets we use a Wide ResNet-28-2 \cite{zagoruyko2016wide} with 1.5M parameters. This architecture makes the dimension of our feature vectors ${d=128}$. Even though we obtain competitive results on many of the splits, we do perform worse than SOTA, especially in the very-low label regime.

\subsubsection{CIFAR-100}

CIFAR-100 is similar to CIFAR-10 in that it consists of color images of size $32\times32$ with training and test sets of size 50,000 and 10,000 respectively. However, CIFAR-100 contains 100 classes, making it a much more challenging classification problem. For this dataset, we use the Wide ResNet-28-8 architecture with 24M parameters, resulting in $d=512$. On this dataset, we achieve SOTA results across all splits, not only beating FixMatch and ReMixMatch, but also the more recent methods DP-SSL and Dash.

\subsubsection{STL-10}

The dataset STL-10 comprises color images of size $96\times96$ belonging to ten classes. It has a labeled training set of 5,000 images and a unlabeled training set of 100,000 images. Its distribution of unlabeled data is wider than the labeled distribution, meaning that the unlabeled set contains classes that are not present in the labeled set. Here we use a Wide ResNet-37-2 with 6M parameters, making $d=256$. On this dataset, we achieve a very competitive error rate, surpassed only by the result reported by Dash.

\subsection{Training speed}

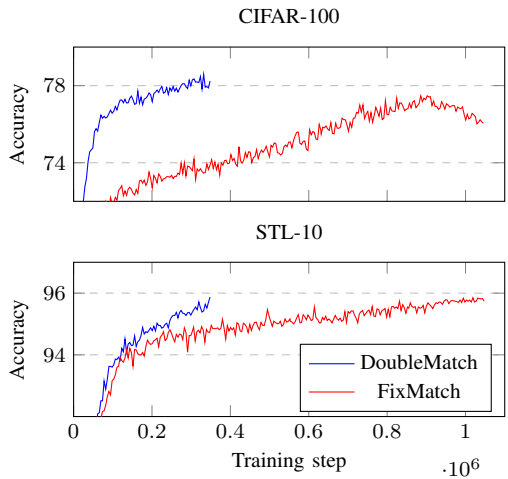
\begin{figure}[]
    \centering
    \footnotesize
    \begin{tikzpicture}
    \begin{groupplot}[
        group style={
            group size=1 by 2,
            x descriptions at=edge bottom,
            vertical sep=23pt,
        },
    ]
    \nextgroupplot[
        title={CIFAR-100},
        width=0.4\textwidth,
        height=0.2\textwidth,
        ylabel={Accuracy},
        xmin=0, xmax=1100000,
        ymin=72, ymax=80,
        legend pos=south east,
        ymajorgrids=true,
        ytick={74,78},
        grid style=dashed,
    ]
        \addplot[color=blue,]
        table[col sep=comma,x expr=\thisrow{Step}/64, y=Value] {figure-data/doublematch-cifar100-10000-acc.csv};
    
        \addplot[color=red,]
        table[col sep=comma,x expr=\thisrow{Step}/64, y=Value] {figure-data/fixmatch-cifar100-10000-acc.csv};

    \nextgroupplot[
        title={STL-10},
        width=0.4\textwidth,
        height=0.2\textwidth,
        xlabel={Training step},
        ylabel={Accuracy},
        xmin=0, xmax=1100000,
        ymin=92, ymax=97,
        ytick={94,96},
        legend pos=south east,
        ymajorgrids=true,
        grid style=dashed
    ]

        \addplot[color=blue,]
        table[col sep=comma,x expr=\thisrow{Step}/64, y=Value] {figure-data/doublematch-stl10-1000-acc.csv};
    
        \addplot[color=red,]
        table[col sep=comma,x expr=\thisrow{Step}/64, y=Value] {figure-data/fixmatch-stl10-1000-acc.csv};

        \legend{DoubleMatch, FixMatch}
        
    \end{groupplot}
\end{tikzpicture}
    \caption{Comparing FixMatch to DoubleMatch on training runs on CIFAR-100 and STL-10. We see that DoubleMatch reaches higher accuracies with roughly one third of the training steps.}
    \label{fig:training-speed}
    \vspace*{-10pt}
\end{figure}

Running FixMatch for its full training duration on CIFAR-100 using a single A100 GPU takes 5 days of wall-time. DoubleMatch reduces these long training times by more efficiently making use of unlabeled data through the added self-supervised loss. While the methods we use as comparison in Table \ref{tab:accuracies} run for little more than 1M training steps, we run DoubleMatch for only roughly a third of that. A clear illustration of our increase in training speed is seen in Fig. \ref{fig:training-speed} where we compare DoubleMatch to FixMatch during training runs on CIFAR-100 with 10,000 labeled training data and on STL-10 with 1,000 labeled data.

\subsection{Discussion}

We present an extension of FixMatch, which outperforms the original work in terms of prediction accuracy and training efficiency on CIFAR-100, STL-10, and 2/3 splits of SVHN. We note a particularly high performance on CIFAR-100 (new SOTA) and STL-10. Our hypothesis for the results on these two datasets is that the self-supervised loss contributes to the biggest improvement when it is difficult to reach high classification accuracies on the unlabeled training set. Low accuracies on the unlabeled set could be either due to 1) the difficulty of the classification problem, as in the case with CIFAR-100, or 2) due to the unlabeled data coming from a wider distribution than the labeled training set, as in the case with STL-10. For CIFAR-10 and SVHN, it is generally possible to reach high accuracies on the unlabeled training set so the quality of the pseudo-labels can be very high, making a self-supervised loss less important.

In Table \ref{tab:accuracies}, we include Dash \cite{xu2021dash} and DP-SSL \cite{xu2021dp}. These are recent methods that, similarly to us, aim to improve the FixMatch framework. However, contrary to DoubleMatch, these methods focus on improving the pseudo-labeling aspect: Dash through a dynamic confidence threshold and DP-SSL by an ensemble of labeling functions. Both of these methods report improvements compared to FixMatch. A future line of work could be combining these improved methods for pseudo-label selections with the self-supervised loss from DoubleMatch.

\subsection{Hyperparameters} \label{sec:hyper}

For CIFAR-10, SVHN and STL-10 we use $\gamma=7/8$ and $w_d=0.0005$. For CIFAR-100 we use $\gamma=5/8$ and $w_d=0.001$. For $w_s$, we use $10, 5, 2$ for CIFAR-100 with 10,000, 2,500 and 400 labels respectively; $5, 1, 0.5$ for CIFAR-10 with 4,000, 250 and 40 labels; $0.05, 0.05, 0.001$ for SVHN with 1,000, 250 and 40 labels; and $1.0$ for STL-10 with 1,000 labels. We use $\eta_0=0.3$, $\mu=7$, $B=64$, $\tau=0.95$ and SGD momentum 0.9 for all datasets. We run all experiments for 352,000 training steps.

\section{Ablation}

In this section we cover two ablation studies related to our method. First, we show results from experiments where the cosine similarity in our self-supervised loss is replaced by other similarity functions. The second ablation study regards the importance of the pseudo-labeling loss for different sizes of the labeled training set.

\subsection{Self-supervised loss functions}

In DoubleMatch we use a cosine similarity for the self-supervised loss term according to \eqref{eq:cosine}. We have conducted experiments with other loss functions. One alternative to the cosine similarity is a simple mean squared error, as used in, e.g.,\ \cite{laine2016temporal}:
\begin{equation}
    l_s = \frac{1}{\mu B d} \sum_{i=1}^{\mu B} (h(v_i) - z_i)^T(h(v_i) - z_i).
    \label{eq:mse}
\end{equation}
Another alternative, as done in, e.g.,\ \cite{caron2021emerging} is to apply the softmax function to the feature vectors and then calculate the cross-entropy loss as
\begin{equation}
    l_s = \frac{1}{\mu B} \sum_{i=1}^{\mu B} H(\sigma(h(v_i)), \sigma(z_i/\lambda)).
\end{equation}
Here, $\lambda$ is a parameter that can be used to sharpen the resulting distribution for the prediction of weakly augmented data. The standard softmax function, $\sigma(\cdot)$, is given by
\begin{equation}
    \sigma_k(x) = \frac{e^{x_k}}{\sum_{j=1}^D e^{x_j}} \quad \text{for} \quad x\in \mathbb{R}^{D \times 1},
\end{equation}
where $\sigma_k$ and $x_k$ are the k:th elements of $\sigma$ and $x$ respectively.

We have evaluated the different loss function on CIFAR-100 with 10,000 labels, where we made training runs using 1) MSE 2) Softmax with $\lambda=1$ and 3) Softmax with $\lambda = 0.1$. The self-supervised weight, $w_s$, is re-tuned for every loss function. The results are shown in Table \ref{tab:loss-functions}. When comparing to the other functions, we note that a considerably lower error rate is reached using the cosine similarity, indicating that this is indeed the correct choice for DoubleMatch.

\begin{table}[]
    \footnotesize
    \centering
    \caption{Evaluations of different functions for the self-supervised loss in DoubleMatch on CIFAR-100 with 10,000 labels.}
    \label{tab:loss-functions}
    \begin{tabular}{c c c }
\toprule
Loss function & Error rate & $w_s$ \\ \midrule
Cosine & 21.22\spm{0.17} & 10.0 \\
MSE & 23.91 & 0.25 \\
Softmax ($\lambda=1$) & 23.23 & 1.0 \\
Softmax ($\lambda=0.1$) & 23.57 & 0.5 \\ \bottomrule
\end{tabular}
\end{table}

\subsection{Importance of pseudo-labels}

Lastly, we analyze the importance of pseudo-labels in DoubleMatch for different sizes of the labeled training set. These experiments are conducted on CIFAR-100. Here, we evaluate the difference in accuracy between DoubleMatch with and without the pseudo-labeling loss, ($l_p$ in \eqref{eq:pseudo-label-loss}). The weight for the self-supervised loss, $w_s$, is re-tuned for each split after removing $l_p$. The drop in accuracy by removing $l_p$ for different numbers of labeled training data is shown in Table \ref{tab:pseudo-importance}. We note that there is nearly no loss in performance by removing $l_p$ when using 10,000 labels. However, when moving towards fewer labels, this reduction in accuracy increases monotonously to roughly 8.5 for 400 labels.

While DoubleMatch still reaches SOTA results for CIFAR-100 with 400 labels, these findings seem to be in line with our poor results on CIFAR-10 and SVHN with 40 labels, indicating that our added self-supervised loss does not do well on its own in the low-label regime. It is also consistent with results reported from Dash \cite{xu2021dash} and DP-SSL \cite{xu2021dp} indicating that improved pseudo-label selections seem to be more important in the low-label regime. However, with enough labels, 
the pseudo-label loss can be completely replaced by a self-supervised loss with little to no loss in performance.


\begin{table}[]
    \footnotesize
    \centering
    \caption{Reduction in test accuracy on CIFAR-100 by removing the pseudo-label loss from DoubleMatch for different sizes of the labeled training set.}
    \label{tab:pseudo-importance}
    \begin{tabular}{c c c c c}
\toprule
& \multicolumn{4}{c}{Nr. of labeled training data} \\ \cmidrule(lr){2-5}
& 400 & 1,000 & 2,500 & 10,000 \\ \midrule
Reduction & 8.46 & 3.81 & 2.20 & 0.39 \\ \bottomrule
\end{tabular}
\end{table}

\section{Conclusion}

This paper shows that using a self-supervised loss in semi-supervised learning can lead to reduced training times and increased test accuracies for multiple datasets. In particular, we present new SOTA results on CIFAR-100 using different sizes of the labeled training set while using fewer training steps than existing methods. Additionally, we present interesting findings showing that, with enough labeled training data, the pseudo-labeling loss can be 
removed with no performance reduction in the presence of a self-supervised loss.

\nocite{larsson-cvpr-2019}


\section*{Acknowledgment}

This work was supported by Saab AB, the Swedish Foundation for Strategic Research, and Wallenberg AI, Autonomous Systems and Software Program (WASP) funded by the Knut and Alice Wallenberg Foundation.
The experiments were enabled by resources provided by the Swedish National Infrastructure for Computing (SNIC) at Chalmers Centre for Computational Science and Engineering (C3SE), and National Supercomputer Centre (NSC) at Linköping University.



\bibliographystyle{IEEEtran}
\bibliography{IEEEabrv,mybib.bib}
%

\end{document}